\documentclass[11pt]{article}

\usepackage[colorlinks,urlcolor=blue,linkcolor=blue,citecolor=blue]{hyperref}
\usepackage{color,array}
\usepackage{graphicx}
\graphicspath{{./figures/}}

\usepackage{amsmath,amssymb}

\usepackage[displaymath,mathlines]{lineno}

\usepackage{enumerate}

\begin{document}

\title{Anisotropic, Sparse and Interpretable Physics-Informed Neural Networks for PDEs}


\author{{Amuthan A. Ramabathiran, Prabhu Ramachandran}
\footnote{Author names listed alphabetically. Both authors contributed equally to this work.}\\ Department of Aerospace Engineering, \&\\ Center for Machine Intelligence and Data Science (C-MInDS),\\ Indian Institute of Technology Bombay, Mumbai 400076, Maharashtra, India.}

\maketitle

\begin{abstract}
There has been a growing interest in the use of Deep Neural Networks (DNNs) to solve Partial Differential Equations (PDEs). Despite the promise that such approaches hold, there are various aspects where they could be improved. Two such  shortcomings are (i) their computational inefficiency relative to classical numerical methods, and (ii) the non-interpretability of a trained DNN model. In this work we present ASPINN, an anisotropic extension of our earlier work called SPINN--Sparse, Physics-informed, and Interpretable Neural Networks--to solve PDEs that addresses both these issues. ASPINNs generalize radial basis function networks. We demonstrate using a variety of examples involving elliptic and hyperbolic PDEs that the special architecture we propose is more efficient than generic DNNs, while at the same time being directly interpretable. Further, they improve upon the SPINN models we proposed earlier in that fewer nodes are require to capture the solution using ASPINN than using SPINN, thanks to the anisotropy of the local zones of influence of each node. The interpretability of ASPINN translates to a ready visualization of their weights and biases, thereby yielding more insight into the nature of the trained model. This in turn provides a systematic procedure to improve the architecture based on the quality of the computed solution. ASPINNs thus serve as an effective bridge between classical numerical algorithms and modern DNN based methods to solve PDEs. In the process, we also streamline the training of ASPINNs into a form that is closer to that of supervised learning algorithms.
\end{abstract}



\section{Introduction}
Learning solutions of Partial Differential Equations (PDEs) using Deep Neural Networks (DNNs) has attracted significant attention over the past few years--see for instance \cite{BN2018}, Physics-Informed Neural Networks (PINNs) \cite{RPK2019} and the Deep Ritz method \cite{EYu2018}. The central idea in these methods is the direct use of either the PDE or its associated variational form to construct the loss function; the origin of this idea goes back to the seminal contributions in \cite{LLF97}. However, these methods remain computationally inefficient in comparison to traditional algorithms like the finite element method. Another drawback of this approach which is not discussed as often is the fact that the trained DNN model is not \emph{interpretable} in the same sense as traditional numerical methods. For instance, the parameters of a finite element approximation can be directly interpreted as the value of the field variables at specified nodal locations in the domain; this interpretation then permits the systematic development of various refinement strategies that improve the approximation in critical regions where the local errors are high. A corresponding strategy does not exist in the context of DNNs on account of the fact that the weights and biases of the DNN have no simple interpretation in terms of the value of the field variable that is approximated. The greater ramifications of a lack of interpretability of the model is discussed forcefully in the context of high-stakes decision making in \cite{rudin_stop_2019}.

It is therefore pertinent to develop interpretable DNN models that permit the development of more efficient algorithms. In this contribution, we build upon our earlier work~\cite{spinn} on SPINN (Sparse, Physics-informed, and Interpretable Neural Networks) and propose special DNN architectures that are sparse and interpretable. Crucially, we discuss how the specific notion of interpretability that is afforded by these specially crafted DNNs naturally leads to more efficient algorithms. Specifically, we develop an extension of SPINN where the local zone of influence of each node is anisotropic; this leads to a more efficient utilization of the nodes and a better resolution of various local features of the solution. This also provides a natural strategy to choose the architecture of these special networks, in sharp contrast to generic DNN based methods where the choice of network architecture is largely ad-hoc.

The larger idea underlying our approach is the construction of DNNs with special architectures that are interpretable. The use of DNNs with special architecture has been in vogue for a few decades in the approximation theoretic literature on DNNs; examples of this include the pioneering contributions in \cite{Mha93} and more recent works like \cite{Yar17, OPS19}.    A particularly noteworthy work in this regard is \cite{HLXZ2020} where the authors illustrate the relation between piecewise linear finite element approximation and deep ReLU networks by giving an explicit construction of the ReLU DNN corresponding to a given piecewise linear finite element discretization. Though the authors in \cite{HLXZ2020} don't view their work through the lens of interpretability, the sparse DNN that they construct is an example of an interpretable neural network whose weights and biases are directly related to the unknown quantities in the finite element discretization. In our earlier work on SPINN \cite{spinn}, we built upon this idea and introduced a class of sparse neural networks that generalize classical Radial Basis Function (RBF) networks \cite{Buh2000} for solving PDEs. In particular, we demonstrated that certain kinds of classical meshless approximations are exactly representable as particular sparse DNNs. We present here an extension of our previous work by introducing a generalization of SPINN architectures--which we called ASPINN (Anisotropic SPINN)--that permits locally anisotropic approximations of the solution which are more efficient than using regular RBFs. 

We remark here that the reinterpretation of an RBF ansatz as a single layer neural network is quite old; see for instance \cite{BL88} for an early discussion of this connection. Further, the class of RBF networks have good universal approximation properties \cite{PS91}. We note in passing that SPINN architecture that we introduced in \cite{spinn} generalizes RBF networks and includes architectures that cannot be reduced to an RBF too. The extension of SPINN that we present inherits all the properties of SPINN and includes more efficient architectures.

It bears emphasis that the development of the proposed architecture is a natural outcome of the interpretability of the SPINN architecture. As elaborated in \cite{spinn}, SPINN models can be viewed as particles associated with a spherical zone of influence. This interpretation naturally suggests the use of particles with anisotropic zones of influence that are optimally designed based on the local solution landscape. This significantly improves the efficiency of the method by reducing the number of required parameters. In one of the examples discussed later a more than 5-fold reduction in the required number of nodes is seen. This simple idea that underlies the ASPINN architecture will be more precise in the next section.

The outline of the paper is as follows. We introduce the structure of the ASPINN architecture and details of implementation in the next section. We then present a variety of results involving elliptic and hyperbolic PDEs that illustrate the utility of ASPINN, and finally conclude with a discussion of the merits and shortcomings of ASPINN.

\section{Methodology}
We present the key ideas underlying ASPINN by focusing on a (linear/nonlinear) PDE of the form
\begin{equation} \label{eq:pde_generic}
L(u) = f, \text{ in } \Omega, \text{ \& } B(u) = g, \text{ on } \partial \Omega.
\end{equation}
In \eqref{eq:pde_generic}, $L$ is a differential operator, $u:\Omega \subseteq \mathbb{R}^d \to \mathbb{R}$ is the field variable of interest, and $f$ and $g$ are given data on the interior and the boundary, respectively. The presentation given here can be extended to time dependent PDEs in a straightforward manner by treating space and time together, as will be illustrated later.

Among the many classical methods that have been developed to solve PDEs like \eqref{eq:pde_generic}, meshless methods (see \cite{li_liu_mm_review_2002} for example) approximate the unknown solution $u$ using an ansatz of the form
\begin{equation} \label{eq:pde_meshless_approx}
u(x) = \sum_{i=1}^N U_i \; \varphi\left(\frac{\lVert x - X_i \rVert}{h_i}\right),
\end{equation}
where $\varphi$ is a kernel function, $\{X_i\}_{i=1}^N$ are node locations, $\{h_i\}_{i=1}^N$ measure the size of the (spherical) zone of influence of each node, and $\{U_i\}$ are the nodal weights. This ansatz has a very simple interpretation: the unknown field $u$ is obtained as a weighted linear combination of shifted and scaled version of the kernel $\varphi$. As elaborated in our earlier work \cite{spinn}, the ansatz \eqref{eq:pde_meshless_approx} can be exactly represented as a sparse DNN. In short, the ansatz \eqref{eq:pde_meshless_approx} can equivalently be viewed as a radial basis function network where the input $x$ is first transformed using a mesh encoding layer to the feature vectors $\{(x - X_i)/h_i\}_{i=1}^N$, transformed subsequently by a kernel layer, and finally linearly combined to exactly represent \eqref{eq:pde_meshless_approx}.

The starting point of the current work is an anisotropic generalization of the ansatz \eqref{eq:pde_meshless_approx} of the form
\begin{equation} \label{eq:aniso_meshless}
u(x) = \sum_{i=1}^N U_i \varphi\left(\left\lVert\Sigma_i^{-1}(x - X_i)\right\rVert \right).
\end{equation}
In \eqref{eq:aniso_meshless}, the matrices $\{\Sigma_i \in \mathbb{R}^{d \times d}\}_{i=1}^N$ are symmetric and positive definite; they represent the local anisotropy of the zone of influence of each node. For the special case when $\Sigma_i = \text{diag}(h_i, \ldots, h_i) \in \mathbb{R}^{d\times d}$, we recover the isotropic ansatz \eqref{eq:pde_meshless_approx}. The anisotropic ansatz \eqref{eq:aniso_meshless} can be expressed as a sparse DNN in a straightforward manner: the input $x \in \mathbb{R}^d$ is first transformed to a mesh encoding layer using weights $\text{diag}(\Sigma_1^{-1}, \ldots, \Sigma_N^{-1})$ and bias vector $[\Sigma_1^{-1}X_1, \ldots, \Sigma_N^{-1}X_N]$--this is called the mesh encoding layer--and then passed into a kernel layer which generalizes the kernel $\varphi$. The output of the kernel layer is linearly combined to get $u(x)$. It is evident that weights and biases of this network directly correspond to terms in the ansatz \eqref{eq:aniso_meshless}. These networks can further be generalized to include non-RBF kernels too, as developed in detail in our earlier work \cite{spinn}. We call this sparse architecture as Anisotropic SPINN, or ASPINN, for short.

The training of the ASPINN model is carried out by minimizing a loss function that is chosen directly in terms of the PDE \eqref{eq:pde_generic}, as was originally proposed in \cite{LLF97}. For instance, the integral of the squared residue of the PDE is a suitable choice for the loss function:
\begin{equation} \label{eq:loss_pde_sq_res}
\mathcal{L}(u) = \int_\Omega \lVert L(u(x)) - f(x) \rVert^2\,dx + \frac{\alpha}{2}\int_{\partial \Omega} \Vert B(u(x)) - g(x) \rVert^2\,dx.
\end{equation}
The boundary conditions are introduced using a simple penalty approach directly in the loss function \eqref{eq:loss_pde_sq_res}, but other approaches are possible--see for instance \cite{BN2018} and \cite{SS22} more interesting alternatives to incorporating the boundary conditions.

In practice, a discretized version of the loss \eqref{eq:loss_pde_sq_res} is used:
\begin{equation} \label{eq:loss_pde_sq_res_disc}
\mathcal{L}_d(u) = \frac{1}{M} \sum_{k=1}^M \lVert L(u(\xi_k)) - f(\xi_k) \rVert^2 + \frac{\alpha}{2\tilde{M}}\sum_{k=1}^{\tilde{M}} \lVert B(u(\tilde\xi_k)) - g(\tilde\xi_k) \rVert^2.
\end{equation}
In \eqref{eq:loss_pde_sq_res}, the points $\{\xi_k \in \Omega\}_{k=1}^M$ and $\{\tilde\xi_k \in \partial \Omega\}_{k=1}^{\tilde{M}}$ denote sampling points in the domain $\Omega$ and on the boundary $\partial \Omega$, respectively. These can be viewed as collocation points where the loss functional \eqref{eq:loss_pde_sq_res} is evaluated.

We remark that we could equivalently use other loss functions; for instance, the variational integral that is used in the Deep Ritz method \cite{EYu2018} is also a suitable choice for the loss function.

The optimization of the loss function is carried out using a suitable variant of the Stochastic Gradient Descent (SGD) algorithm. In this work, we use the Adam optimization algorithm \cite{kingma2014} for training ASPINN.

The training method presented above is quite standard, but we wish to emphasize two particular features of the approach we use to train ASPINN:
\begin{enumerate}[(i)]
\item In typical applications of PINN to solve PDEs, the loss \eqref{eq:loss_pde_sq_res_disc} is minimized to train the network, but there is no independent testing of the trained model. To make the training process more streamlined with the standard training-testing procedure in supervised learning, we identify separate sets of interior testing points $\{\eta_k \in \Omega\}_{k=1}^{K}$ and boundary testing points $\{\tilde\eta_k \in \partial \Omega\}_{k=1}^{\tilde{K}}$, in addition to the interior training points $\{\xi_k \in \Omega\}_{k=1}^M$ and the boundary training points $\{\tilde\xi_k \in \partial\Omega\}_{k=1}^{\tilde{M}}$. We train the networks on the training samples and test them on the test samples to get a sense of the generalization capabilities of the trained model. We note that this approach is applicable for PINNs too.

\item A naive implementation of the optimization process outlined above does not work for ASPINN since a gradient descent update does not guarantee the symmetry and positive definiteness of the matrices $\{\Sigma_i\}_{i=1}^N$. To enforce this (nonlinear) constraint, we use (a slight modification of) the log-Cholesky parametrization of symmetric and positive definite matrices \cite{PB96}. Specifically, we choose as training parameters the lower triangular matrices $\{\tilde{L}_i \in \mathbb{R}^{d\times d}\}_{i=1}^N$; the entries of $\tilde{L}_i$ are denoted as $(\tilde{l}_{ij})_{i,j=1}^d$. The matrix $\Sigma_i$ is then constructed as
\begin{equation} \label{eq:log_Cholesky_param}
\begin{split}
\Sigma_i &= L_i L_i^T,\\
L_i &= \begin{pmatrix} \exp s\tilde{l}_{11} & 0 & \ldots & 0\\ \tilde{l}_{21} & \exp s\tilde{l}_{22} & \ldots & 0\\ \vdots & \ddots & & \vdots\\ \vdots & & \ddots & \vdots\\ \tilde{l}_{d1} & \tilde{l}_{d2} & \ldots & \exp s\tilde{l}_{dd} \end{pmatrix}
\end{split}
\end{equation}
In \eqref{eq:log_Cholesky_param}, $s$ is a scale parameter that can be used to control the growth of the diagonal entries of $\Sigma$ during optimization. The use of the log-Cholesky parametrization \eqref{eq:log_Cholesky_param} yields an unconstrained  optimization problem for the matrices $\{\Sigma_i\}_{i=1}^N$, and is compatible with gradient descent updates. We remark that even a simple Cholesky decomposition would work here, but it suffers from degeneracy due to the lack of constraint on the diagonal entries of $L$ being positive; we also found that the Cholesky decomposition doesn't perform as well as the log-Cholesky decomposition in practice.

\item Once an ASPINN model is trained, the zone of influence of each node is obtained by computing the eigenvalues of the matrices $\{\Sigma_i\}_{i=1}^N$. For each node $i$, the zone of influence is an ellipsoid; the eigenvalues of the corresponding $\Sigma_i$ dictate the dimensions of the ellipsoid. For vanilla SPINN, the zone of influence is a sphere, which is obtained as a special case of this construction.
\end{enumerate}

\section{Results}
The application of ASPINN to solve various PDEs is now presented to highlight its strengths in comparison to both SPINN and PINN. The ASPINN algorithm was implemented in PyTorch \cite{pytorch}. The simulations were automated using Automan \cite{automan:2018}. The visualizations were prepared using Mayavi \cite{mayavi} and Matplotlib \cite{mpl}. For all the simulations, the Adam optimizer \cite{kingma2014} was used. The scale factor the log-Cholesky parametrization was chosen as $s = 0.5$. For all the simulations, the nodes are initially chosen to have an isotropic zone of influence. The parameter $\alpha$ in the loss function \eqref{eq:loss_pde_sq_res_disc} is chosen to be larger than $\tilde{M}$. This choice is made to ensure that the boundary conditions are satisfied to a reasonable degree of accuracy. The nodes are initially placed uniformly over the domain in all the simulations presented below.

\subsection{2D Poisson equation}
\begin{figure}
\centering
\includegraphics[width=0.8\textwidth]{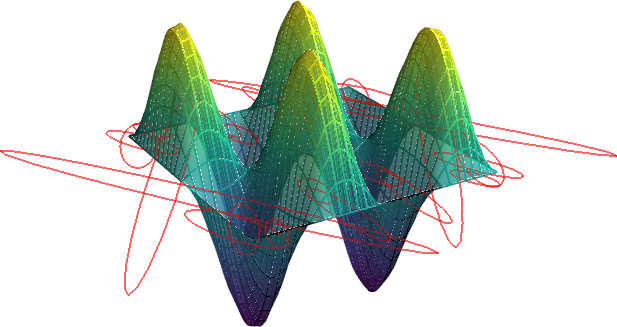}
\caption{Solution of \eqref{eq:poisson_2d} using ASPINN. The exact solution is shown as a wireframe plot.}
\label{fig:poisson_sine_aspinn_soln}
\end{figure}

\begin{figure}
\centering
\includegraphics[width=0.8\textwidth]{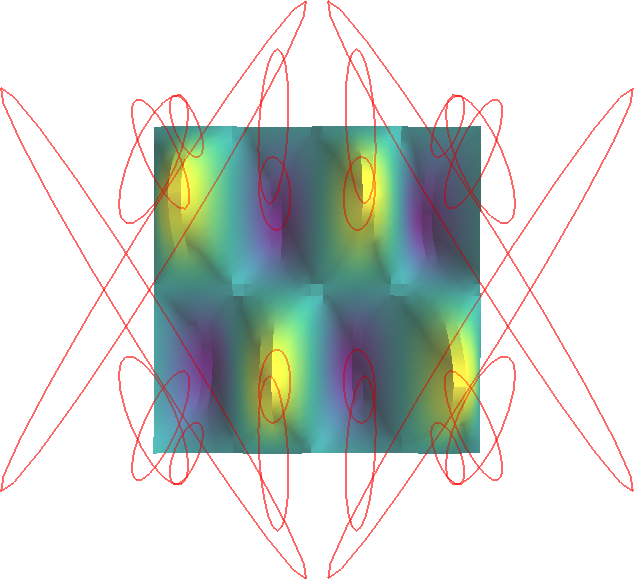}
\caption{Location of nodes and their anisotropic zones of influence for the ASPINN solution of \eqref{eq:poisson_2d}.}
\label{fig:poisson_sine_centers}
\end{figure}

We begin with the Poisson equation in two dimensions:
\begin{equation} \label{eq:poisson_2d}
\begin{split}
-\nabla^2 u(x,y) &= 5\pi^2 \sin (2\pi x) \sin (\pi y),\\
 (x, y) \in \Omega &= [-1,1]\times[-1,1],\\
u(x,y) = 0, &\quad (x,y) \in \partial \Omega.
\end{split}
\end{equation}
The PDE \eqref{eq:poisson_2d} admits the exact solution $u(x,y) = \sin(2\pi x)\sin(\pi y)$. We show a plot of the solution obtained using ASPINN in comparison with the exact solution in Figure~\ref{fig:poisson_sine_aspinn_soln}. The location of the centers along with the zones of influence of each node--ellipses in the 2D case--are shown in Figure~\ref{fig:poisson_sine_centers}. The ASPINN model had 4 nodes along the X direction and 2 along the Y direction, comprising a total of 8 nodes. Around $200$ interior sampling points were chosen to evaluate the loss. The choice of the number of nodes was dictated by the nature of the solution--as can be seen from Figure~\ref{fig:poisson_sine_aspinn_soln}, the exact solution has 8 peak; natural choice of the ASPINN architecture is one where an initially isotropic node is placed uniformly over the domain close to the location of each peak. As is apparent from Figure~\ref{fig:poisson_sine_centers}, the nodes towards their respective peaks, and their zones of influence become appropriately anisotropic so as to best captures the solution locally. For instance, it can be seen that the exact solution has greater variation along the Y direction than the X direction; the zones of influence of the interior nodes accordingly are elongated along the Y direction and contracted along the X direction. The purpose of this simple example is to illustrate the fact that the direct translation of the weights and biases of the ASPINN model into corresponding interpretable quantities like the anisotropy of the zones of influence and the location of the nodes permit us to reason both about the architecture and the quality of the solution. To understand the latter, we recall that SGD algorithms often find a local minimum in the loss landscape and it is not often clear whether the local minimum that is obtained is a good one. In this case however, the reasoning just outlined informs us that the weights and biases are in the \emph{right} local minimum since the corresponding nodes and zones of influence behave as per our expectations of their optimal values ought to be. It bears emphasis that such an analysis about the weights and biases of generic DNNs are not feasible in general. The ability to directly interpret the trained parameters of the ASPINN model in this fashion thus permits us to reason about various architectural choices and quality of the local minima of the trained solution.

\begin{figure}
\centering
\includegraphics[width=0.8\textwidth]{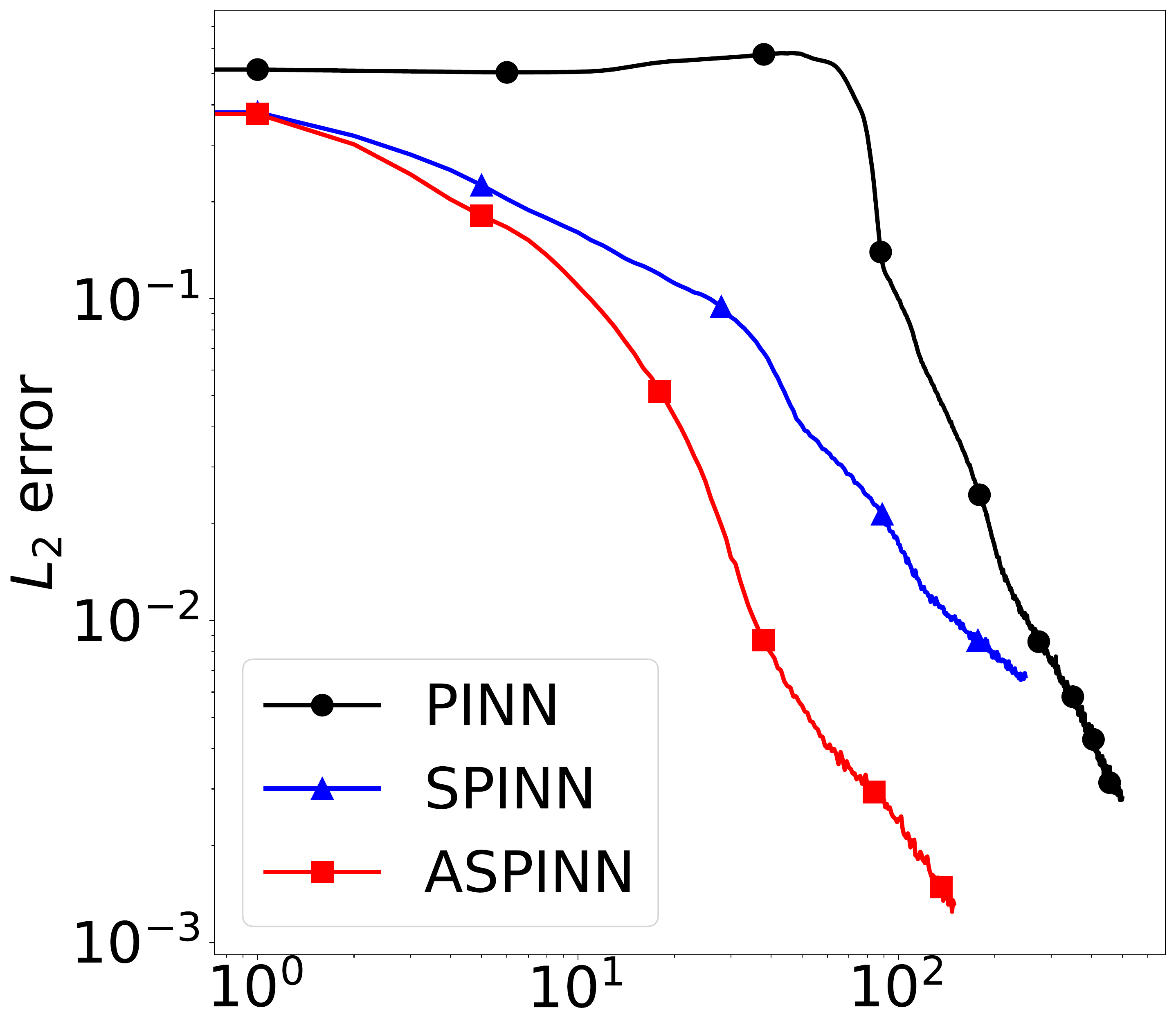}
\caption{$L_2$ error during SGD training for the PDE \eqref{eq:poisson_2d}.}
\label{fig:poisson_sine_comp_3_L2}
\end{figure}

A comparison of the $L_2$ errors of the computed solution using PINN, SPINN and ASPINN is shown in Figure~\ref{fig:poisson_sine_comp_3_L2}. The PINN architecture was chosen to have a comparable number of neurons with respect to the A/SPINN models. The results clearly indicate the superior performance of ASPINN in comparison to both SPINN and PINN. Stated differently, it takes much smaller number of iterations for the ASPINN model to reach a desired level of accuracy compared to PINN. Since the order of accuracy increases with increasing the number of nodes of the ASPINN model, this also implies that fewer nodes are required to learn a solution of comparable accuracy to a given SPINN or PINN model. This observation will be borne out by the various simulations presented in the sequel.

\begin{figure}
\centering
\includegraphics[width=0.8\textwidth]{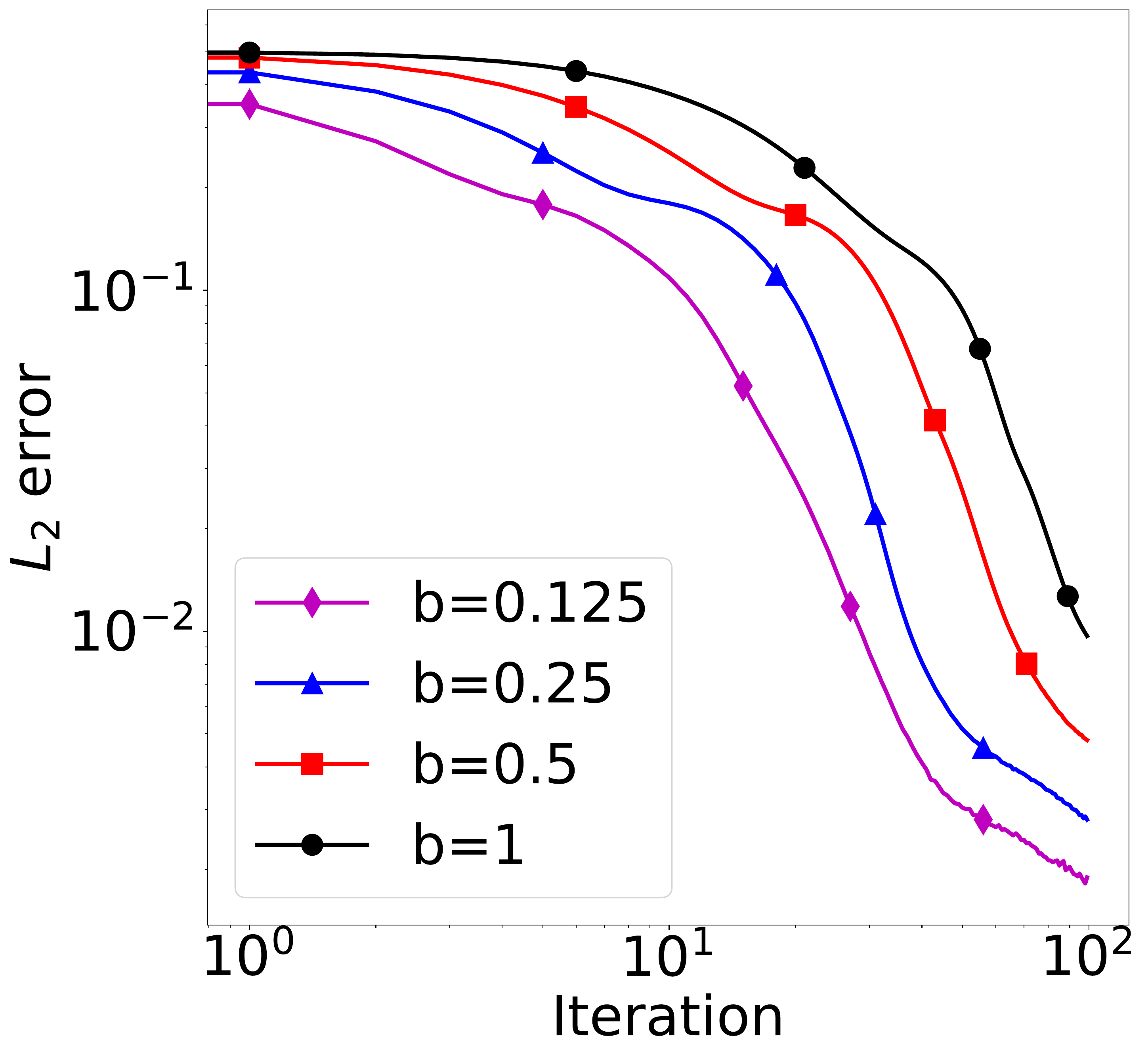}
\caption{Effect of batch size on $L_2$ error for ASPINN. The values indicated in the label refer to the fraction of the sampling points that is used for each batch.}
\label{fig:poisson_sine_batch}
\end{figure}

The effect of batch size on $L_2$ error obtained using ASPINN is shown in Figure~\ref{fig:poisson_sine_batch}. The numerical evidence indicates that smaller batch sizes are in general better for obtaining faster convergence. This observation holds in general when solving other PDEs too.

\subsection{Static ripple in 2D}
As our next example, we choose the PDE
\begin{equation} \label{eq:ripple_2d}
\begin{split}
-\nabla^2 u(x,y) & + 16\pi^2(4x^2 + y^2)u(x,y)\\
 & = 2(x^2 + y^2 - 2)\cos(2\pi(2x^2 + y^2)\\
 & - 4\pi(3(1 - x^2)(1 - y^2)\\
 & - 8x^2(1 - y^2) - 4y^2(1 - x^2))\sin(2\pi(2x^2 + y^2)),\\
 & (x, y) \in \Omega = [-1,1] \times [-1,1],\\
u(x,y) &= 0, \text{ on } \partial \Omega.
\end{split}
\end{equation}
The PDE \eqref{eq:ripple_2d} admits an exact solution $u(x,y) = (1 - x^2)(1 - y^2)\cos(2\pi(2x^2 + y^2))$. The solution looks like a static ripple over the domain $\Omega = [-1,1] \times [-1,1]$. This particular PDE is chosen to illustrate the performance of ASPINN in the presence of large gradients distributed spatially over the domain.

\begin{figure}
\centering
\includegraphics[width=0.8\textwidth]{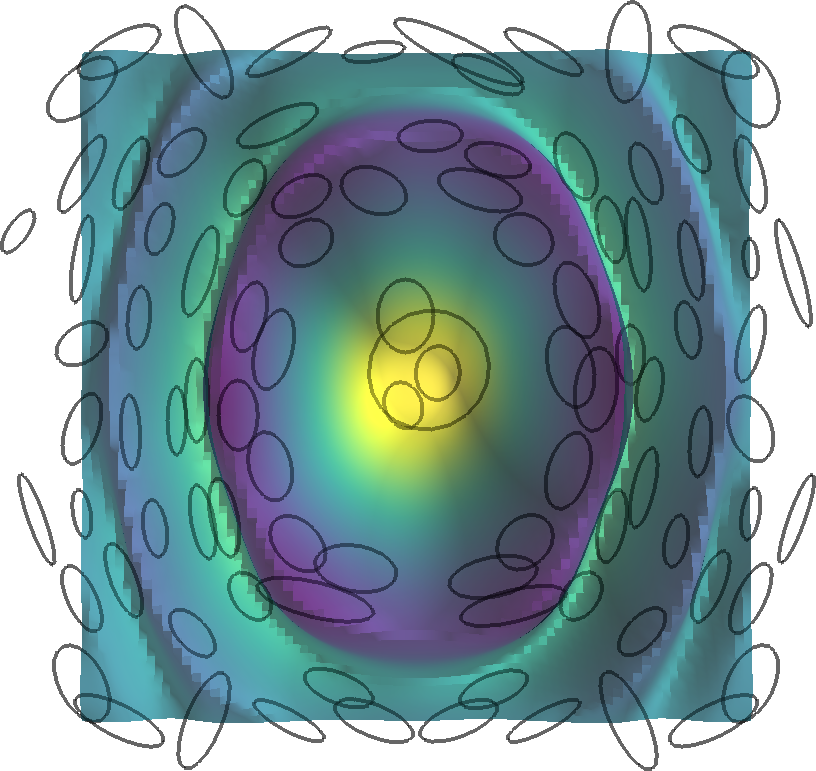}
\caption{Location of nodes and their anisotropic zones of influence for the ASPINN solution of \eqref{eq:ripple_2d}.}
\label{fig:ripple_centers}
\end{figure}

An example of a trained ASPINN model is shown in Figure~\ref{fig:ripple_centers}. The figure clearly illustrates how the nodes adapt their location and orientation during the learning process to capture the various gradients in the exact solution. It is worth emphasizing that this kind of information that directly relates the weights of the trained model to physically interpretable parameters is not possible with the use of generic DNNs as in the case of PINNs.

As mentioned in the previous example, another advantage in having a visual representation of the weights as shown in Figure~\ref{fig:ripple_centers} is that it helps us evaluate the quality of the local minimum that the algorithm lands in. The fact that the centers are not in symmetric positions indicate that the local optimum is not a global optimum; on the other hand, the alignment of the nodes along the solution gradients indicate that the local minimum that the optimizer steers the model to is indeed a good one.

\begin{figure}
\centering
\includegraphics[width=0.8\textwidth]{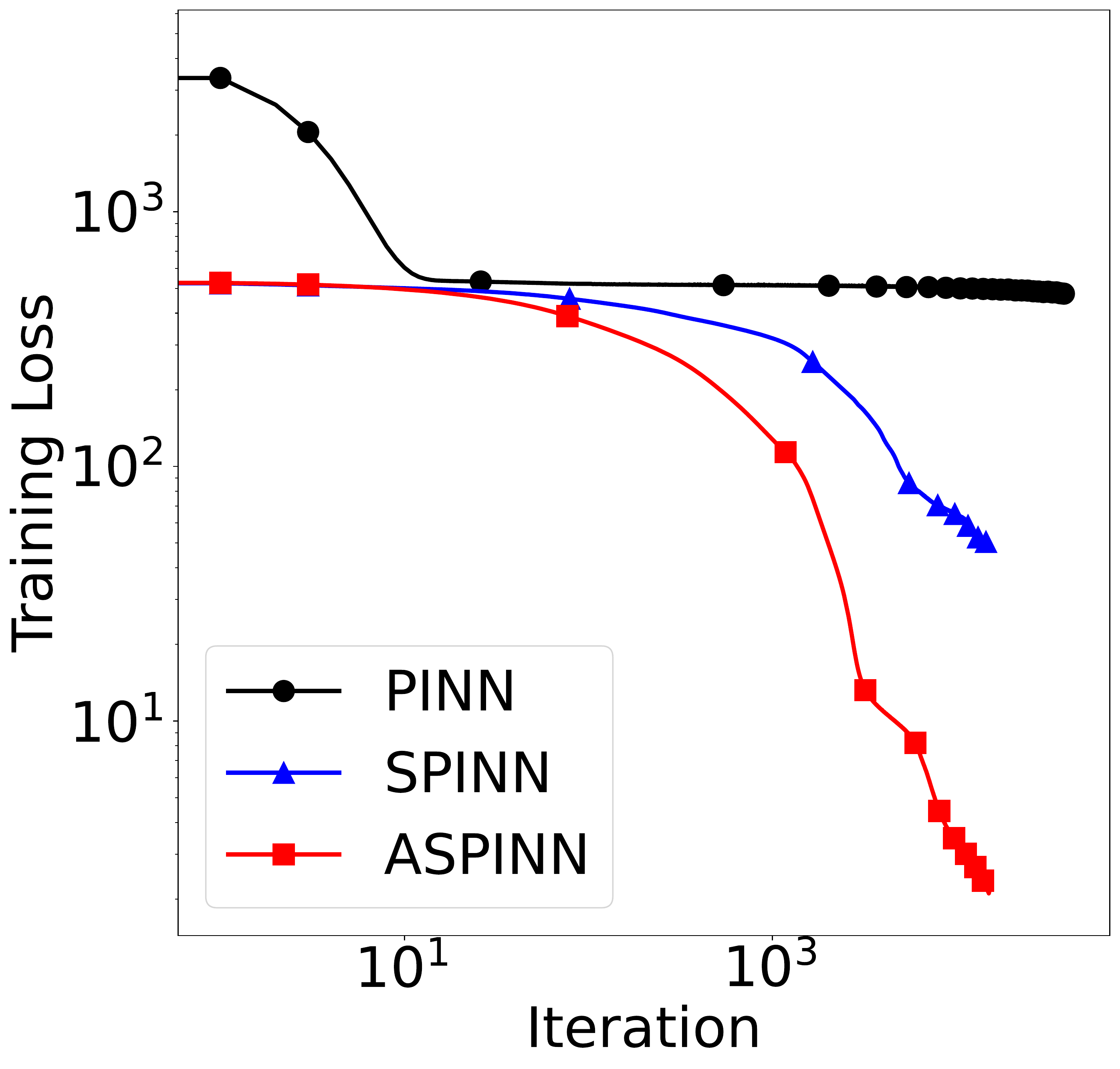}
\caption{Training loss of PINN, SPINN and ASPINN for the PDE \eqref{eq:ripple_2d}. For SPINN and ASPINN, 64 nodes and 1600 sampling points were used. The PINN model was chosen as a 2 layer tanh-network with 100 neurons in each layer. The batch size was chosen as 200.}
\label{fig:ripple_comp}
\end{figure}

The training loss for PINN, SPINN and ASPINN is shown in Figure~\ref{fig:ripple_comp}. This example clearly indicates the advantage of using ASPINN--in addition to being interpretable, it is much more efficient in computing an approximate solution to the PDE.

\begin{figure}
\centering
\includegraphics[width=0.8\textwidth]{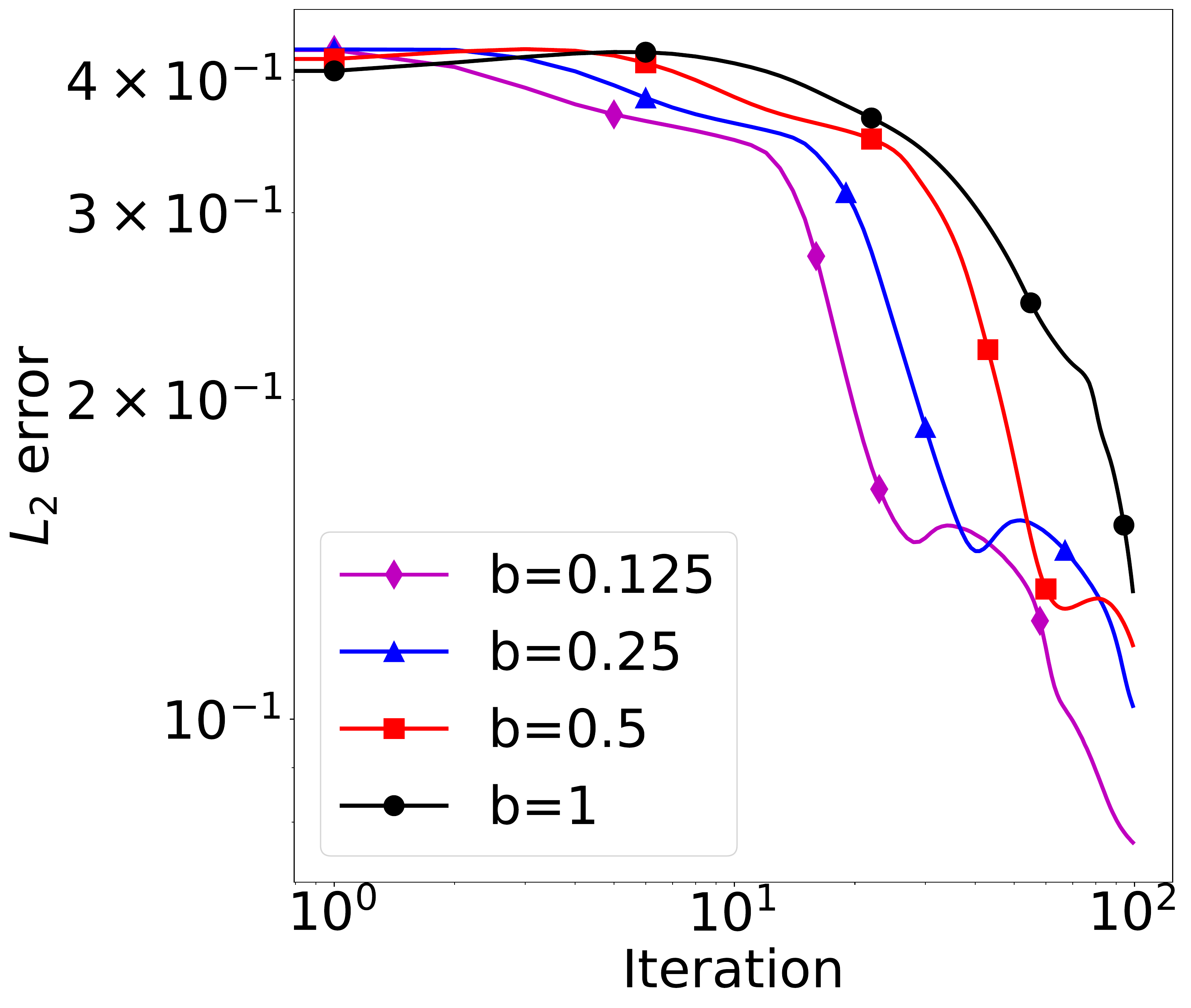}
\caption{Effect of batch size on $L_2$ error for ASPINN. The values indicated in the label refer to the fraction of the sampling points that is used for each batch.}
\label{fig:ripple_batch}
\end{figure}

The effect of batch size on the convergence of the ASPINN model is shown in Figure~\ref{fig:ripple_batch}. As noted earlier, the numerical simulations suggest that smaller batch sizes work better.

\subsection{Square slit}
We consider next a problem involving a non-convex domain, namely the Poisson equation on a square with a slit:
\begin{equation} \label{eq:square_slit}
\begin{split}
\nabla^2 u(x,y) + 1 &= 0, \; (x, y) \in \Omega = (-1,1)\times(-1,1) \setminus [0,1),\\
u(x,y) &= 0 \; (x,y) \in \partial \Omega.
\end{split}
\end{equation}
This equation does not admit an exact solution, but can be solved using any standard numerical method, like the finite element method for instance.

\begin{figure}
\centering
\includegraphics[width=0.8\textwidth]{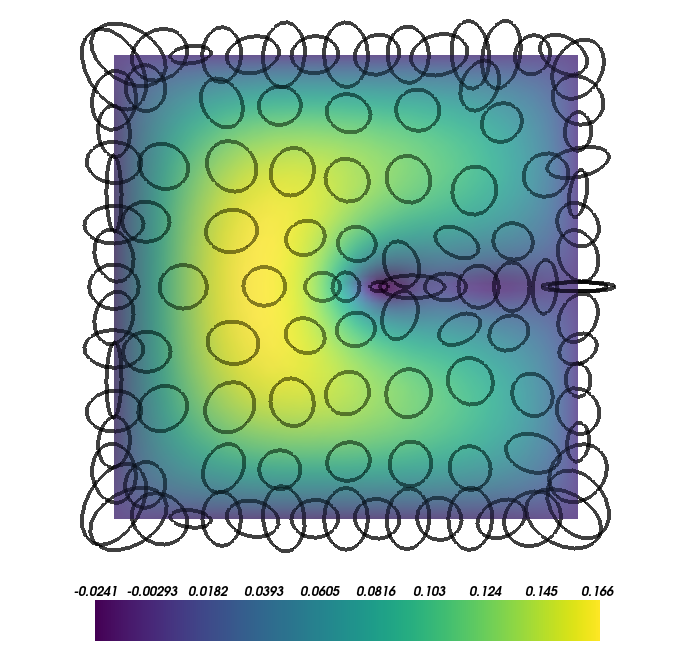}
\caption{Solution of Poisson equation over a square domain with a slit \eqref{eq:square_slit} using spacetime ASPINN. The $L_2$ error of the converged solution with respect to a reference finite element solution is $3.3e-3$.}
\label{fig:square_slit_soln}
\end{figure}

The solution of the PDE \eqref{eq:square_slit} using ASPINN is shown in Figure~\ref{fig:square_slit_soln}. The visualization of the zones of influence illustrates the ability of the SPINN algorithm to adapt to the local solution landscape.

\begin{figure}
\centering
\includegraphics[width=0.8\textwidth]{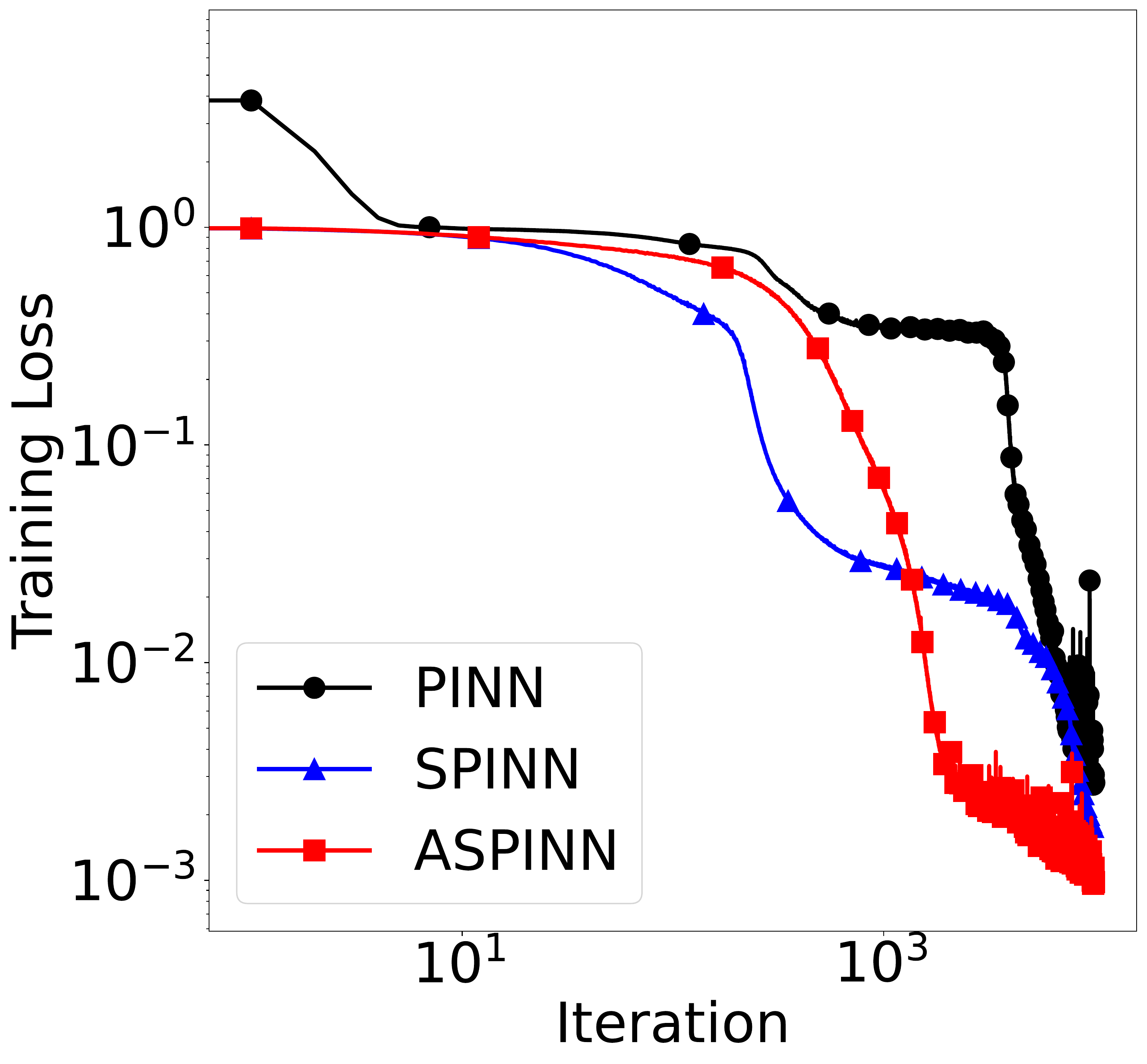}
\caption{Training loss of PINN, SPINN and ASPINN for the PDE \eqref{eq:square_slit}. For SPINN and ASPINN, 49 nodes and 1200 sampling points were used. The PINN model was chosen as a 2 layer tanh-network with 100 neurons in each layer. The batch size was chosen as 32.}
\label{fig:square_slit_comp}
\end{figure}

A comparison of the training loss for PINN, SPINN and ASPINN for this PDE is shown in Figure~\ref{fig:square_slit_comp}. It can be seen that ASPINN outperforms both SPINN and PINN in this case. It is to be remarked that this specific example was chosen since both PINN and SPINN exhibit poor solution convergence. Though the accuracy of the ASPINN model is not high--the $L_2$ error computed with reference to a finite element solution is of the order of $10^{-3}$ and plateaus there--the speed of convergence of the ASPINN algorithm is much faster than PINN and SPINN.

\subsection{1D linear advection equation}
We now turn our attention to time dependent PDEs. We focus on hyperbolic PDEs since the presence of a propagating wave yields sharp gradients in the spacetime solution, thereby serving as good tests of the ASPINN models.

The linear advection equation takes the form
\begin{equation} \label{eq:lin_adv_1d}
\begin{split}
\frac{\partial u(x,t)}{\partial t} + a \frac{\partial u(x,t)}{\partial x} &= 0, \; x \in \mathbb{R}, t \in [0,T],\\
u(x,0) &= u_0(x), \; \lim_{x \to \pm \infty} u(x,t) = 0.
\end{split}
\end{equation}
The linear advection equation \eqref{eq:lin_adv_1d} admits traveling wave solutions of the form $u(x,t) = f(x - at)$. We choose in particular a Gaussian initial profile of the form $u_0(x) = \exp(-(x - \mu)^2/2\sigma^2)$ with $\mu = -0.3$ and $\sigma = 0.15$.

\begin{figure}
\centering
\includegraphics[width=0.8\textwidth]{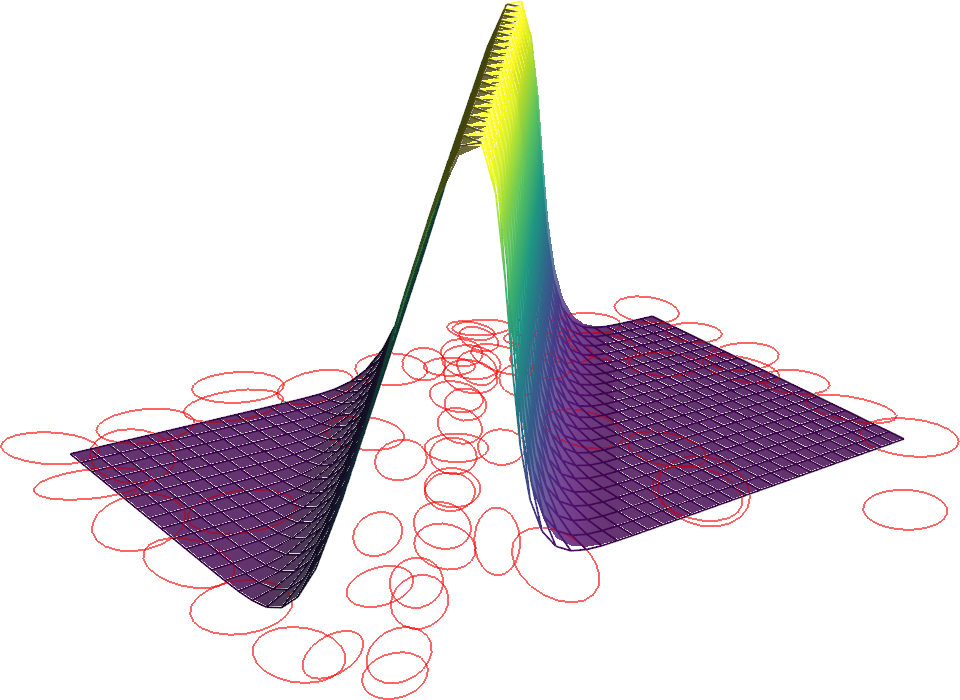}
\caption{Solution of advection equation \eqref{eq:lin_adv_1d} using spacetime ASPINN.}
\label{fig:advection_soln}
\end{figure}

\begin{figure}
\centering
\includegraphics[width=0.8\textwidth]{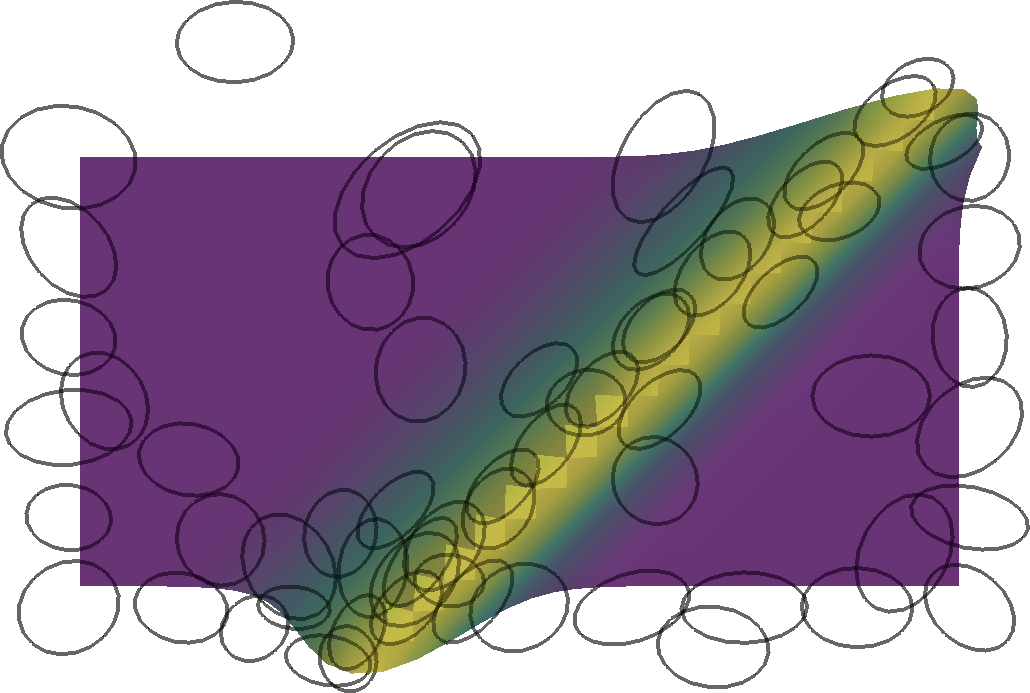}
\caption{Location of nodes for spacetime ASPINN solution of the advection equation \eqref{eq:lin_adv_1d}.}
\label{fig:advection_centers}
\end{figure}

For ease of computing the solution, we focus on the interval $[-1,1]$ and simulate the solution over the interval $[0,T]$, with $T$ chosen such that the wave does not exit the interval $[-1,1]$ during this time period. The solution obtained using ASPINN is shown in Figure~\ref{fig:advection_soln}. The solution was computed by simultaneously discretizing space and time; we call this spacetime ASPINN. The location of the corresponding nodes is shown in Figure~\ref{fig:advection_centers}. The orientation of the nodes along with their zones of influence along the propagating wave gives a good illustration of the effectiveness and interpretability of the ASPINN algorithm. This further suggests that for PDEs whose solutions have known structural properties, the weights and biases of ASPINN models can be chosen appropriately to accelerate convergence. For instance, a placement of initially isotropic nodes along the propagating wave would yield much faster convergence. Such a problem-specific initialization of the weights and biases is not possible in general for PINNs owing to their non-interpretability.

\begin{figure}
\centering
\includegraphics[width=0.8\textwidth]{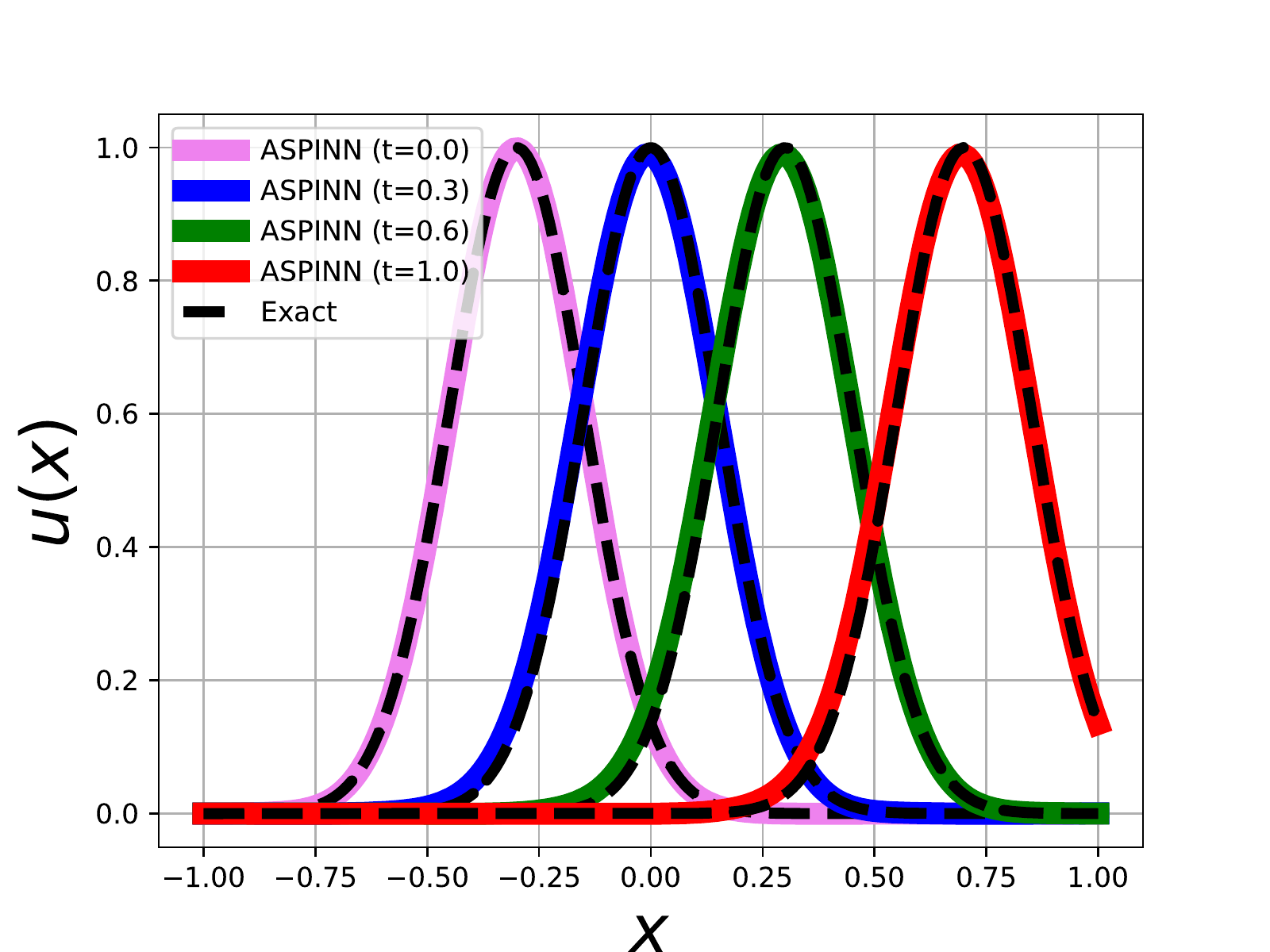}
\caption{A comparison of the spacetime solution of the linear advection equation \eqref{eq:lin_adv_1d} computed using ASPINN with 40 internal nodes and the exact traveling wave solution with an initial Gaussian profile.}
\label{fig:lin_adv_soln}
\end{figure}

\begin{figure}
\centering
\includegraphics[width=0.8\textwidth]{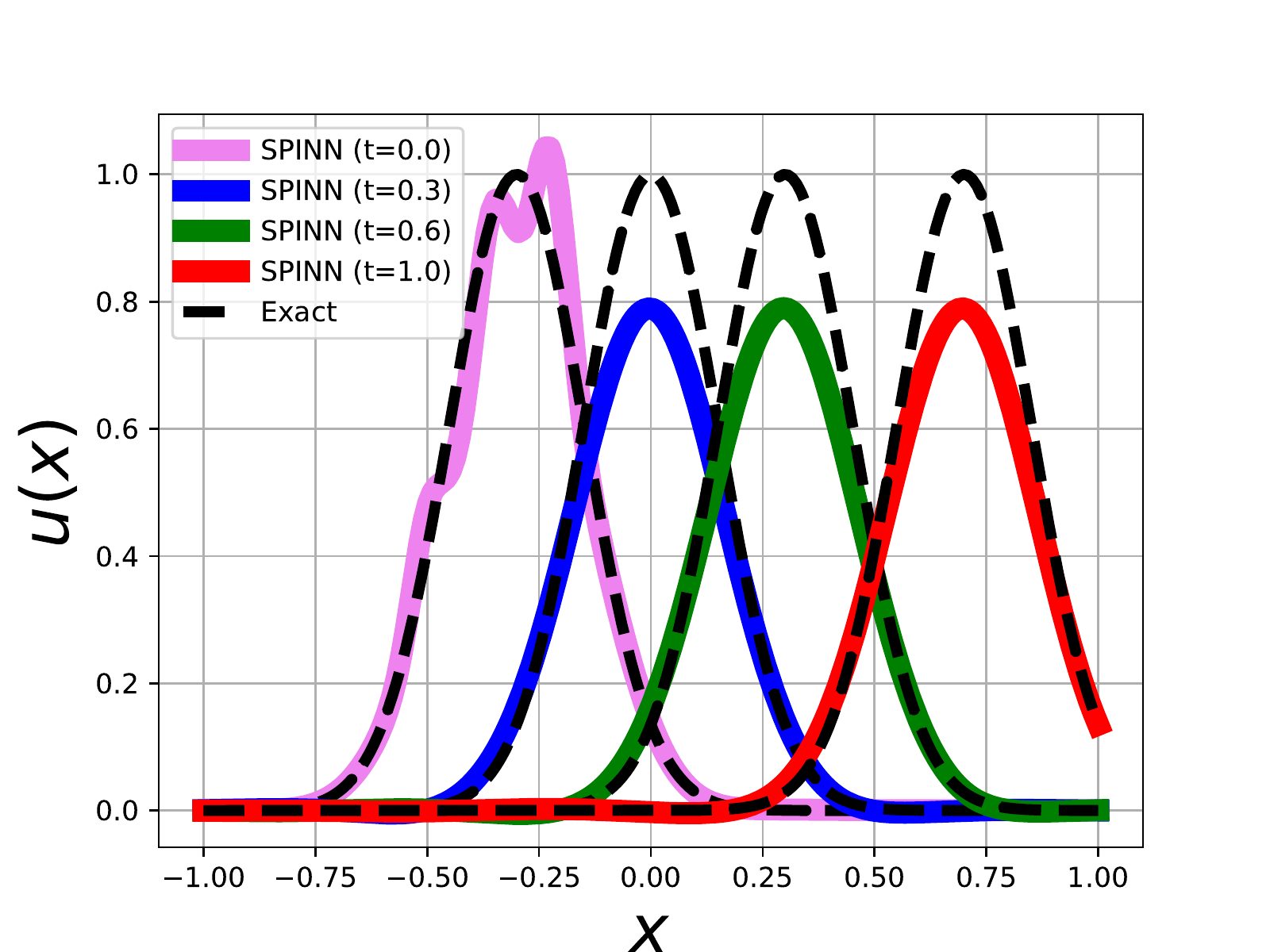}
\caption{A comparison of the spacetime solution of the linear advection equation \eqref{eq:lin_adv_1d} computed using SPINN with 40 internal nodes and the exact traveling wave solution with an initial Gaussian profile.}
\label{fig:lin_adv_soln_spinn}
\end{figure}

Finally, various time snapshots of the spacetime ASPINN solution using 40 internal nodes were extracted and explicitly compared with the exact solution, as shown in Figure~\ref{fig:lin_adv_soln}. It can be seen that ASPINN captures the traveling wave profile without any numerical diffusion, thereby proving the effectiveness of the ASPINN algorithm for studying hyperbolic PDEs.  For reference we also show the solution obtained using the isotropic SPINN using the same number of nodes (40) in Figure~\ref{fig:lin_adv_soln_spinn}. The solution of the ASPINN method is much better than that of the SPINN method.

\subsection{1D Burgers equation}
As a final example we discuss the inviscid Burgers equation:
\begin{equation} \label{eq:burgers_1d}
\begin{split}
\frac{\partial u(x,t)}{\partial t} + u(x,t) \frac{\partial u(x,t)}{\partial x} &= 0, \; x \in (-1,1), t \in [0,T],\\
u(x,0) &= \begin{cases} \sin 2\pi(x + \frac{1}{2}), & -\frac{1}{2} \le x \le \frac{1}{2}\\ 0, &\text{otherwise}.\end{cases}\\
u(-1,t) = u(1,t) &= 0, \; t \in [0,T].
\end{split}
\end{equation}
The reason for choosing the inviscid Burgers equation is that the solution develops a shock after a finite time. In traditional numerical treatments of the Burgers equation, special care has to be taken to capture the shock accurately.

\begin{figure}
\centering
\includegraphics[width=0.8\textwidth]{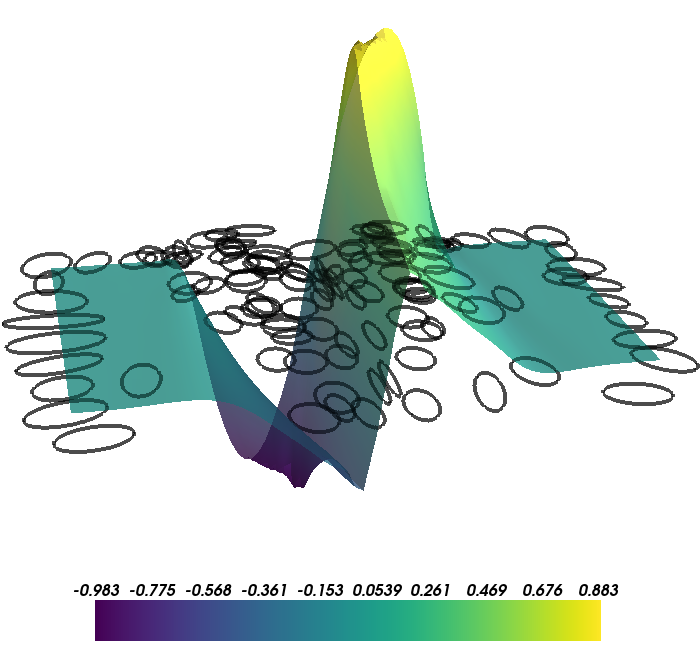}
\caption{Solution of Burgers equation \eqref{eq:burgers_1d} using spacetime ASPINN. The ASPINN model has 80 interior nodes, 600 interior samples, and batch size 100.}
\label{fig:burgers_soln}
\end{figure}

\begin{figure}
\centering
\includegraphics[width=0.8\textwidth]{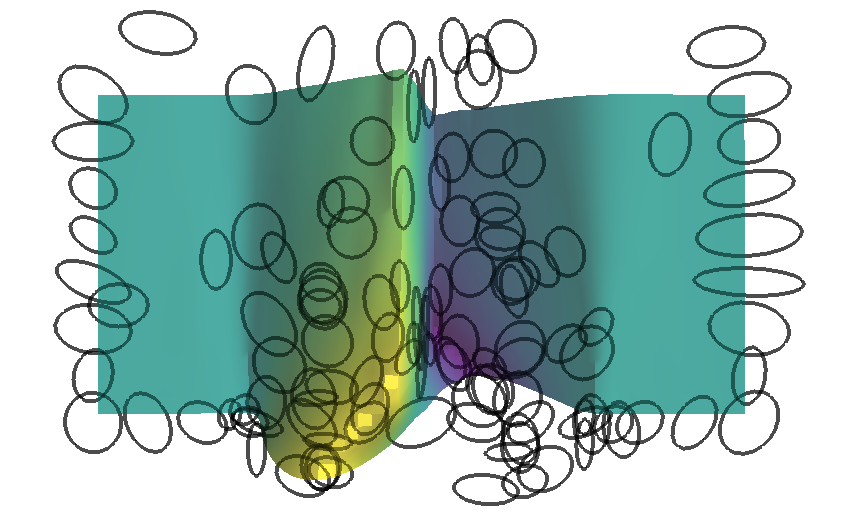}
\caption{Location of nodes for spacetime ASPINN solution of the Burgers equation \eqref{eq:burgers_1d}.}
\label{fig:burgers_centers}
\end{figure}

We adopt a spacetime discretization to solve the Burgers equation, just as in the case of linear advection. The spacetime solution computed using ASPINN is shown in Figure~\ref{fig:burgers_soln}. The location of the nodes and their zones of influence is shown in Figure~\ref{fig:burgers_centers}. It can be seen clearly from Figure~\ref{fig:burgers_centers} that the zones of influence follow the shock and get more elongated along the time direction as they get closer to the shock, which is what would be expected from an adaptive algorithm. It bears emphasis that it is the interpretability of the ASPINN model that permits us to reason about the quality of the computed solution.

\begin{figure}
\centering
\includegraphics[width=0.8\textwidth]{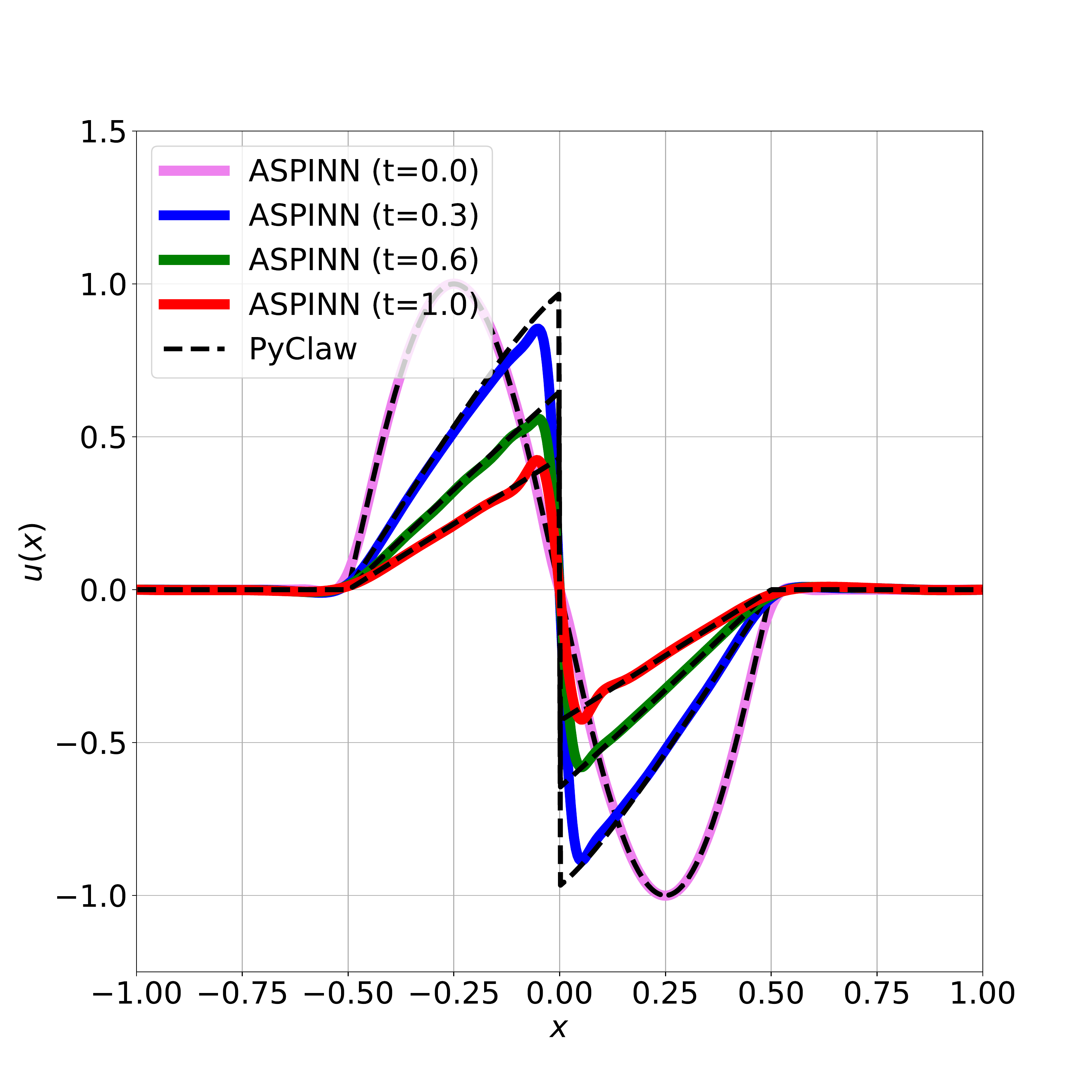}
\caption{A comparison of the spacetime solution of the Burgers equation \eqref{eq:burgers_1d} computed using ASPINN with 80 internal nodes and a reference solution computed using PyClaw \cite{pyclaw}.}
\label{fig:burgers_comp_pyclaw}
\end{figure}

\begin{figure}
\centering
\includegraphics[width=0.8\textwidth]{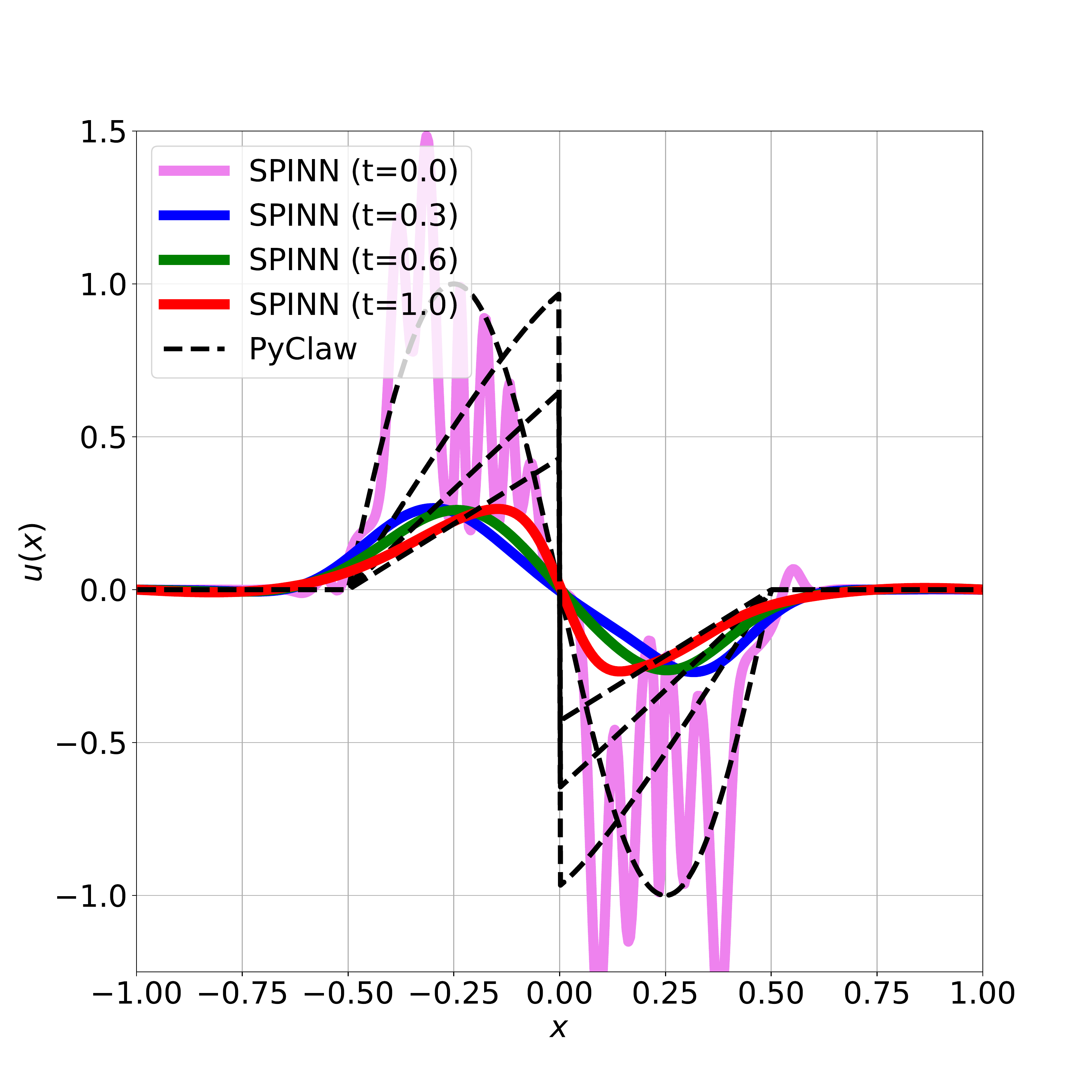}
\caption{A comparison of the spacetime solution of the Burgers equation \eqref{eq:burgers_1d} computed using SPINN with 80 internal nodes and a reference solution computed using PyClaw \cite{pyclaw}.}
\label{fig:burgers_comp_pyclaw_spinn}
\end{figure}

As in the case of linear advection, various time snapshots of the spacetime ASPINN and a reference solution computed using PyClaw \cite{pyclaw} are shown in Figure~\ref{fig:burgers_comp_pyclaw}. For reference we also show the solution using the isotropic SPINN with the same number of internal nodes in Figure~\ref{fig:burgers_comp_pyclaw_spinn}. The ASPINN solution captures the shock very well, there is a small reduction in the peak at $t=0.3$ but subsequently the solution matches very well.  In terms of computational effort, for the same number of nodes and samples the original SPINN algorithm took around 370 seconds while the ASPINN model took 493 seconds to execute on an NVIDIA GeForce RTX 3050 GPU with an AMD Ryzen 7 4800H using PyTorch version 1.12.1.  This is a 30\% reduction in performance. However, it bears emphasis that with the same number of nodes, the isotropic SPINN model produces unusable solutions as can be clearly seen in Figure~\ref{fig:burgers_comp_pyclaw_spinn}. Obtaining accurate results with SPINN would require least 5 times as many nodes to capture with the original isotropic SPINN model. The results presented in \cite{spinn} used around 430 nodes for a similar accuracy.

\section{Discussion}
The examples presented thus far illustrate the effectiveness of the anisotropic extension of SPINN that we propose in this work. The results clearly indicate that ASPINN provides a more efficient and interpretable alternative to PINN. We would now like to discuss a few related issues and point out potential limitations of ASPINN.

\begin{enumerate}[(i)]
\item The sense in which ASPINN is interpretable requires clarification. The notion of \emph{interpretability} is interpreted differently by different researchers. In the context of ASPINN, we refer to interpretability in the sense of understanding the computational graph. This is quite distinct from the emphasis in fields like interpretable/explainable AI--see for instance Locally Interpretable Model-Agnostic Explanations (LIME) \cite{RSG16} where the goal is to find a local fit using an interpretable model to the predictions of a generic DNN model. In contrast, in ASPINN, we design a network with a special architecture such that the weights and biases of this network are interpretable in a physically relevant sense.

\item The interpretability of the weights and biases of an ASPINN model yields useful insights into the refinement of the architecture. For instance, a large error in a particular region immediately suggests adding more nodes and/or sampling points in that region--this translates to an easily implementable change in the architecture. In contrast, when using PINNs, the choice of architecture is largely ad-hoc. Furthermore, there is no systematic means to get a better architecture if the solution displays high local errors.

\item The approach to creating interpretable models that we adopt here is that of designing special architectures. This, however, does not address the issue of the interpretability of generic DNN models. As remarked in \cite{spinn}, the best we can do is to view a DNN as a (global) Ritz approximation; the DNN model learns global basis functions which are then linearly combined. But such an interpretation lacks the sharpness of interpretations of special architectures like SPINN or the ReLU-FEM networks developed in \cite{HLXZ2020}. The development of methods to interpret generic DNN models remains an open challenge.

\item In addition to being interpretable, we emphasize that special architectures like the ones we propose here overcome some of the disadvantages of using a dense DNN to represent the solution. The sparsity of the architecture often translates to a corresponding increase in computational efficiency, though these too remain inefficient in comparison to traditional algorithms like the finite element method. The proposed method thus lies in the spectrum between traditional solvers and DNN based methods, thereby acting as a bridge between different modeling viewpoints.

\item ASPINN can equivalently be viewed as a generalization of classical meshless methods in a manner that renders them differentiable end-to-end. This allows for the use of ASPINN models in conjunction with a larger PDE constrained optimization problem--we will be exploring this in a future work.

\item A fundamental limitation associated with methods like ASPINN is an exact handling of boundary conditions. The present article adopts the simplest means to implement boundary conditions, namely by penalizing deviations from the prescribed boundary conditions. While this is certainly the simplest, it is not the best possible option. A variety of alternatives like the ones discussed in \cite{BN2018, SS22} can be adopted in conjunction with ASPINN.

\item While the examples presented here are limited to simple domains, the method presented is general enough to handle complex domains too. A few such examples were presented in our earlier work \cite{spinn}; since ASPINN follows the larger structure of SPINN, its extension to problems defined on complex boundaries is straightforward. In addition to this, hybrid methods like finite differencing in time in conjunction with ASPINN in space, which were explored in our earlier work with SPINN \cite{spinn}, are also easily implementable with ASPINN. We will be exploring this in a future work, especially in the context of the Burgers equation where capturing the shock is a non-trivial challenge for spacetime discretizations.

\item The approach presented in this work streamlines the training and testing of PINN/SPINN/ASPINN models by choosing distinct sets of sampling points for training and testing. This is necessary to study the generalization capabilities of the trained models.

\item An advantage that is often claimed for neural networks is their ability to overcome the curse of dimensionality. While this claim is not unconditionally true, a question that we haven't discussed in this work is how well special architectures like SPINN/ASPINN fare when it comes to overcoming the curse of dimensionality. Using the arguments presented in works like \cite{Pet22, PS91}, SPINN/ASPINN are expected to be at least as good as RBFs in terms of their approximation capabilities. How well the more general ASPINN models overcome the curse of dimensionality remains to be explored.

\item As a final remark, we note that a variety of extensions of ASPINN, like the use of a kernel network to model the kernel, the use of methods like Least Squares Gradient Descent \cite{CGPPT20} to accelerate the performance, etc. can be easily applied to SPINN too. These and similar extensions are planned for future works.
\end{enumerate}

\section{Conclusion}

The present work provides a general and easy-to-implement numerical framework, called ASPINN, to solve PDEs. ASPINNs generalize classical RBFs by performing a local coordinate transformation for each node in a manner that best captures the solution at the location of the node. This provides a straightforward interpretation of the weights and biases of the network. Such an interpretation helps in developing a systematic means to modify the network's architecture. This removes many of the ad-hoc architectural choices that are made when using generic DNNs, as in PINN or the Deep Ritz method. ASPINNs thus inherit the advantages of DNN based approaches to solve PDEs while retaining strong connections to classical RBF approximations.  Further, the fact that weights and biases of the network can be visualized in a manner that yields useful insight into the nature of the trained model is of special significance to developing adaptive solution strategies. In practice, this translates to ASPINN models requiring much fewer neurons to achieve a solution of desired accuracy compared to generic DNNs, thereby permitting the study of much larger PDEs. The relocation of the nodes to physically interesting regions of the solution and the reorientation of the zones of influence to align with the local solution landscape illustrates the adaptivity of our algorithm, and further allow us to reason about the quality of the trained ASPINN model. ASPINN can be used to solve linear/nonlinear and static/time-dependent PDEs defined on complex domains. Indeed, we presented a variety of results illustrating the efficiency and interpretability of ASPINN in the context of both elliptic and hyperbolic PDEs; our results indicate that ASPINN is in general more efficient than PINNs. Unlike classical approximation schemes, ASPINN is end-to-end differentiable, which opens up new avenues for PDE constrained optimization problems. ASPINN thus provides an efficient, interpretable and differentiable generalization of classical RBF meshless methods, thereby providing a link between classical and modern numerical methods to solve PDEs.

\bibliographystyle{plain}
\bibliography{references}

\begin{thebibliography}{10}

\bibitem{BN2018}
Jens Berg and Kaj Nyström.
\newblock A unified deep artificial neural network approach to partial
  differential equations in complex geometries.
\newblock 317:28--41.

\bibitem{BL88}
D.S. Broomhead and D.~Lowe.
\newblock Multivariable functional interpolation and adaptive networks.
\newblock {\em Complex Systems}, pages 321--355, 1988.

\bibitem{Buh2000}
M.D. Buhmann.
\newblock Radial basis functions.
\newblock {\em Acta Numerica}, pages 1--38, 2000.

\bibitem{CGPPT20}
Eric~C. Cyr, Mamikon~A. Gulian, Ravi~G. Patel, Mauro Perego, and Nathaniel~A.
  Trask.
\newblock Robust training and initialization of deep neural networks: {A}n
  adaptive basis viewpoint.
\newblock In Jianfeng Lu and Rachel Ward, editors, {\em Proceedings of The
  First Mathematical and Scientific Machine Learning Conference}, volume 107 of
  {\em Proceedings of Machine Learning Research}, pages 512--536, Princeton
  University, Princeton, NJ, USA, 20--24 Jul 2020. PMLR.

\bibitem{EYu2018}
Weinan E and Bing Yu.
\newblock The deep ritz method: A deep learning-based numerical algorithm for
  solving variational problems.
\newblock {\em Commun. Math. Stat.}, 6:1--12, 2018.

\bibitem{HLXZ2020}
Juncai He, Lin Li, Jinchao Xu, and Chunyue Zheng.
\newblock {ReLU} deep neural networks and linear finite elements.
\newblock {\em Journal of Computational Mathematics}, 38(3):502--527, 2020.

\bibitem{mpl}
J.~D. Hunter.
\newblock Matplotlib: A 2d graphics environment.
\newblock {\em Computing in Science \& Engineering}, 9(3):90--95, 2007.

\bibitem{pyclaw}
David~I. Ketcheson, Kyle~T. Mandli, Aron~J. Ahmadia, Amal Alghamdi, Manuel
  {Quezada de Luna}, Matteo Parsani, Matthew~G. Knepley, and Matthew Emmett.
\newblock {PyClaw: Accessible, Extensible, Scalable Tools for Wave Propagation
  Problems}.
\newblock {\em SIAM Journal on Scientific Computing}, 34(4):C210--C231,
  November 2012.

\bibitem{kingma2014}
Diederik~P. Kingma and Jimmy Ba.
\newblock Adam: A method for stochastic optimization, 2014.
\newblock Published as a conference paper at the 3rd International Conference
  for Learning Representations, San Diego, 2015.

\bibitem{LLF97}
I.~E. {Lagaris}, A.~{Likas}, and D.~I. {Fotiadis}.
\newblock Artificial neural networks for solving ordinary and partial
  differential equations.
\newblock {\em IEEE Transactions on Neural Networks}, 9(5):987--1000, 1998.

\bibitem{li_liu_mm_review_2002}
Shaofan Li and Wing~Kam Liu.
\newblock Meshfree and particle methods and their applications.
\newblock {\em Applied Mechanics Reviews}, 55(1):1--34, January 2002.
\newblock Publisher: American Society of Mechanical Engineers Digital
  Collection.

\bibitem{Mha93}
H.N. Mhaskar.
\newblock Approximation properties of a multilayered feedforward artificial
  neural network.
\newblock {\em Adv. Comput. Math.}, 1:61--80, 1993.

\bibitem{OPS19}
J.~A.~A. Opschoor, P.~C. Petersen, and Ch. Schwab.
\newblock Deep {ReLU} networks and high-order finite element methods.
\newblock Technical Report 2019-07, Seminar for Applied Mathematics, ETH
  Z{\"u}rich, Switzerland, 2019.

\bibitem{PS91}
J.~Park and I.W. Sandberg.
\newblock Universal approximation using radial-basis-function networks.
\newblock {\em Neural Computation}, 3:246--257, 1991.

\bibitem{pytorch}
Adam Paszke, Sam Gross, Francisco Massa, Adam Lerer, James Bradbury, Gregory
  Chanan, Trevor Killeen, Zeming Lin, Natalia Gimelshein, Luca Antiga, Alban
  Desmaison, Andreas Kopf, Edward Yang, Zachary DeVito, Martin Raison, Alykhan
  Tejani, Sasank Chilamkurthy, Benoit Steiner, Lu~Fang, Junjie Bai, and Soumith
  Chintala.
\newblock Pytorch: An imperative style, high-performance deep learning library.
\newblock In H.~Wallach, H.~Larochelle, A.~Beygelzimer, F.~d\textquotesingle
  Alch\'{e}-Buc, E.~Fox, and R.~Garnett, editors, {\em Advances in Neural
  Information Processing Systems 32}, pages 8024--8035. Curran Associates,
  Inc., 2019.

\bibitem{peng_xai_2022}
Xi~Peng, Yunfan Li, Ivor~W. Tsang, Hongyuan Zhu, Jiancheng Lv, and Joey~Tianyi
  Zhou.
\newblock {XAI} {Beyond} {Classification}: {Interpretable} {Neural}
  {Clustering}.
\newblock {\em Journal of Machine Learning Research}, 23(6):1--28, 2022.

\bibitem{Pet22}
P.C. Petersen.
\newblock Neural network theory, 2022.

\bibitem{PB96}
J.C. Pinheiro and D.M. Bates.
\newblock Unconstrained parametrizations for variance-covariance matrices.
\newblock {\em Statistics and Computing}, 6:289--296, 1996.

\bibitem{RPK2019}
Maziar Raissi, Paris Perdikaris, and George~E Karniadakis.
\newblock Physics-informed neural networks: A deep learning framework for
  solving forward and inverse problems involving nonlinear partial differential
  equations.
\newblock {\em Journal of Computational Physics}, 378:686--707, 2019.

\bibitem{spinn}
Amuthan~A. Ramabathiran and Prabhu Ramachandran.
\newblock {SPINN}: Sparse, physics-based, and partially interpretable neural
  networks for {PDE}s.
\newblock {\em Journal of Computational Physics}, 110600, 2021.

\bibitem{automan:2018}
Prabhu Ramachandran.
\newblock automan: A python-based automation framework for numerical computing.
\newblock {\em Computing in Science \& Engineering}, 20(5):81--97, Sep./Oct.
  2018.

\bibitem{mayavi}
Prabhu Ramachandran and Ga\"{e}l Varoquaux.
\newblock Mayavi: {3D} visualization of scientific data.
\newblock {\em Computing in Science and Engineering}, 13(2):40--51, 2011.

\bibitem{RSG16}
M.T. Ribiero, S.~Singh, and C.~Guestrin.
\newblock ``{Why} should {I} trust you?'' {E}xplaining the predictions of any
  classifier, 2016.
\newblock KDD 16: Proceedings of the 22nd ACM SIGKDD Interational Conference on
  Konwledge Discovery and Data.

\bibitem{rudin_stop_2019}
Cynthia Rudin.
\newblock Stop explaining black box machine learning models for high stakes
  decisions and use interpretable models instead.
\newblock {\em Nature Machine Intelligence}, 1(5):206--215, May 2019.

\bibitem{SS22}
S.~Sukumar and A.~Srivastava.
\newblock Exact imposition of boundary conditions with distance functions in
  physics-informed deep neural networks.
\newblock {\em Computer Methods in Applied Mechanics and Engineering},
  389:114333, 2022.

\bibitem{Yar17}
D.~Yarotsky.
\newblock Error bounds for approximations with deep relu networks.
\newblock {\em Neural Networks}, 94:103--114, 2017.

\end{thebibliography}
\end{document}